\documentclass[sn-mathphys]{sn-jnl}

\usepackage{multirow}
\usepackage{threeparttable}
\usepackage{makecell}
\usepackage{amsmath,bm}
\usepackage{makecell}
\usepackage{bbding}
\usepackage{amssymb}
\usepackage{pifont}
\usepackage{fancyhdr} 
\usepackage{hyperref}

\jyear{2021}%

\theoremstyle{thmstyleone}%
\theoremstyle{thmstyletwo}%

\theoremstyle{thmstylethree}%

\begin{document}

\fancyhf{}
\fancyhead[R]{DOI:xxxxxxxxx}

\title[]{Prompt-Guided Primitive Alignment for Compositional Temporal Grounding}

\author[1]{\fnm{Jianhua} \sur{Yang}}

\author[2]{\fnm{Keji} \sur{He}}

\author[3]{\fnm{Junhui} \sur{Yin}}

\author[1]{\fnm{Yan} \sur{Huang}}

\author[1]{\fnm{Liang} \sur{Wang}}

\affil[1]{\orgdiv{New Laboratory of Pattern Recognition (NLPR), State Key Laboratory of
 Multimodal Artificial Intelligence Systems (MAIS)}, \orgname{ Institute of Automation, Chinese Academy of Sciences (CASIA)}, \orgaddress{\city{Beijing} \postcode{100190},  \country{China}}}

\affil[2]{\orgdiv{School of Artificial Intelligence}, \orgname{Shandong University (SDU)}, \orgaddress{\city{Jinan} \postcode{‌250100‌},  \country{China}}}

\affil[3]{\orgdiv{School of Artificial Intelligence}, \orgname{University of Science and Technology Beijing (USTB)}, \orgaddress{\city{Beijing} \postcode{‌100083},  \country{China}}}


\abstract{Compositional Temporal Grounding (CTG) aims to localize specific moments in untrimmed videos based on sentence queries containing novel compositions of known concepts, posing a significant challenge for robust vision-and-language understanding. Existing methods often exhibit coarse alignment between textual primitives and target moments and lack rich cross-modal interactions between video and language features. To overcome these limitations, we propose a Prompt-guided Primitive Alignment (PPA) framework for CTG. PPA incorporates a Bidirectional Masked Feature Modeling (BMFM) strategy that enhances cross-modal interactions by jointly reconstructing masked video tokens from textual cues and masked word tokens from visual cues via a co-attention mechanism. Furthermore, a Primitive Prompt Feature Modulation (PPFM) module is introduced to explicitly align textual primitives with target video regions through learnable Gaussian weighting, thereby improving primitive-to-moment correspondence and compositional generalization. Extensive experiments on Charades-CG and ActivityNet-CG benchmarks demonstrate that PPA achieves state-of-the-art performance with strong generalization capability to novel compositions and unseen words.}




\maketitle
\section{Introduction}
\label{sec1}

\begin{figure*}[!tb]
\centering
\includegraphics[width=0.9\linewidth]{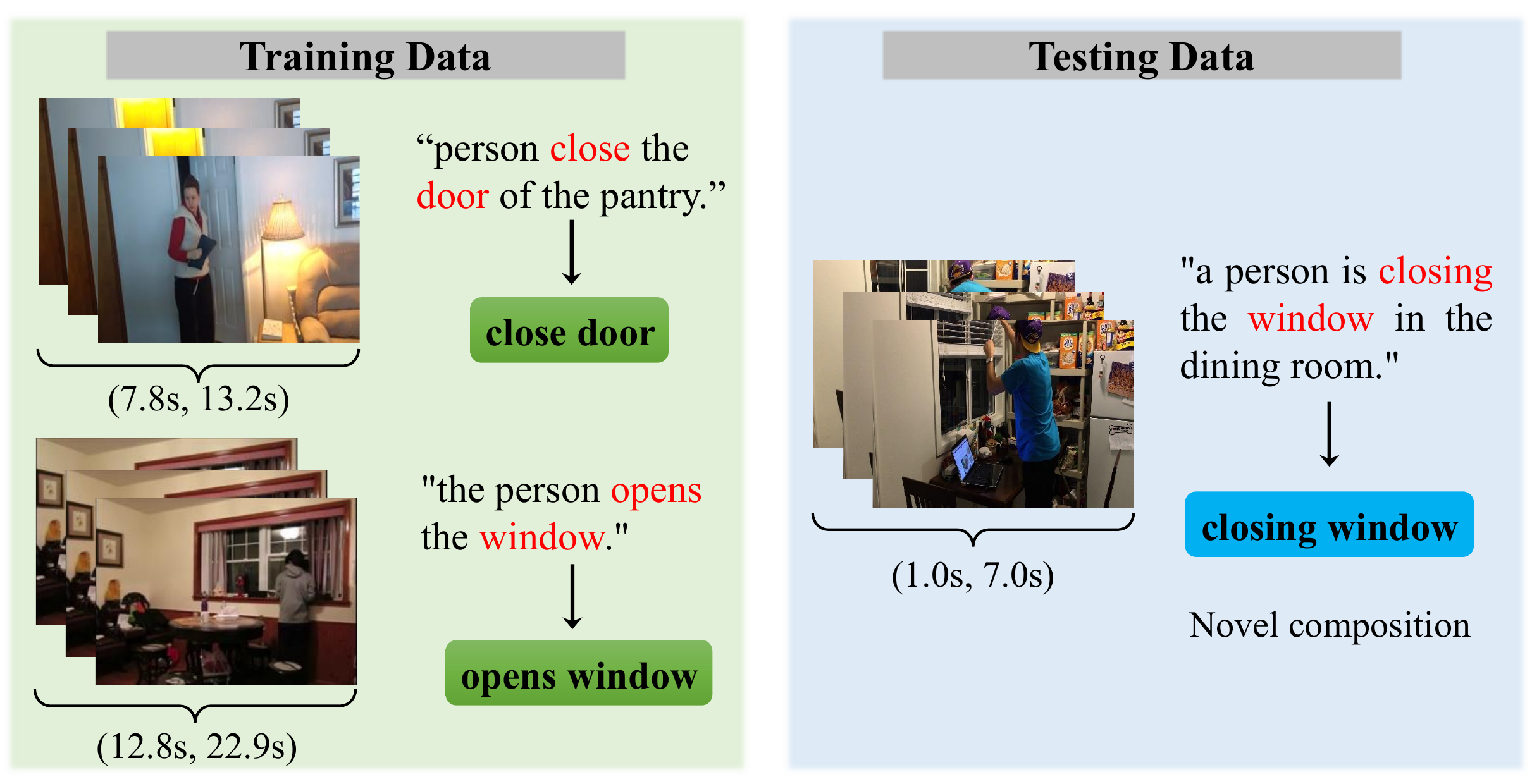}
\caption{\textbf{Illustration of the compositional temporal grounding (CTG)}. The goal of the task is to localize the referred video moment in an untrimmed video described by a given sentence, where the sentence contains novel compositions in testing samples. For example, the testing sample containing the novel composition of ``closing window'', such a novel composition does not exist in the training data, while its components (\emph{i.e.}, ``close'' and ``window'') exist in the training data.}
\label{fig:figure0}
\vspace{-0.2cm}
\end{figure*}

Understanding new inputs by recombining learned concepts is a fundamental aspect of human cognition. Similar to human language that makes systematic generalizations from limited data \cite{lake2018generalization_sys}, the compositional generalization capability is critical in artificial intelligence systems. However, recent studies in the vision-and-language (V\&L) domain \cite{pantazopoulos2022combine,singh2023coarse,trager2023linear} have shown that existing V\&L models exhibit notable limitations in handling unseen compositions of known concepts, as evidenced by their insensitivity to word order and the semantic structure of primitive elements.

Video temporal grounding (VTG) is a fundamental V\&L task that aims to localize a specific moment within an untrimmed video according to its semantic correspondence with the sentence query \cite{gao2017tall,anne2017localizing}. More recently, increasing attention has been devoted to investigating compositional generalization in VTG task, commonly referred to as compositional temporal grounding (CTG) \cite{visa_cvpr2022,sesi_cvpr2023,deco_iccv2023,shine_ceev2024,mesm_aaai2024}. Beyond addressing the conventional VTG task, CTG further investigates a model’s capacity to generalize to queries with novel compositions of seen words as well as those containing novel words. As shown in Fig. \ref{fig:figure0}, the novel composition of ``\emph{closing window}'' in the testing sample is never seen in the training data, while the individual primitives (\emph{i.e.}, ``\emph{close}'' and ``\emph{window}'') are seen in the training data. If a model has been trained on samples illustrating how to ``\emph{close door}'' and ``\emph{opens window}'', it is expected to be capable of compositional generalization to the novel composition of ``\emph{closing window}''. Although existing VTG approaches \cite{2dtan_2020aaai,vlsnet_2020alc,lgi_cvpr2020,sul2023mr} have achieved impressive localization performance, Li et al. \cite{visa_cvpr2022} and Cheng \cite{shine_ceev2024} demonstrated that they struggle to generalize to queries containing novel compositions of seen words. 

To address this challenging problem, the pioneer work VISA \cite{visa_cvpr2022} decomposes videos and queries into structured hierarchies (\emph{i.e.}, global events, local actions, and atomic objects) and devises a variational cross-graph to model semantic correspondence among these concepts. Later, SESI \cite{sesi_cvpr2023} proposes a self-supervised framework that enhances 2D-TAN's \cite{2dtan_2020aaai} sensitivity to equivariant and invariant samples, which are generated by masking different primitives in videos and queries. DeCo \cite{deco_iccv2023} constructs compositional and negative queries through a decomposition-recombination strategy, and establishes multi-granularity correspondence between videos and queries with contrastive ranking loss. In contrast, SHINE \cite{shine_ceev2024} generates hierarchically hard negative samples with large language models (\emph{e.g.}, GPT-3.5-Turbo \cite{gpt_nips2020}), and introduces a saliency-aware two-stage ranking strategy to model multi-granularity semantic relationship between video and hard negatives.  

Although these approaches have made encouraging progress in the CTG task, two critical limitations hinder their further improvements in compositional generalization. \textit{First, weak correspondence between textual primitives and the target moment.} Most existing methods primarily establish semantic alignment at a coarse sentence or phrase level, without explicitly modeling how primitive elements (\emph{e.g.}, verbs and nouns) govern the temporal boundaries of the target moment. Such alignment often overlooks the fact that verbs dictate the precise start and end times, while background entities frequently appear throughout the video, resulting in diluted attention and inaccurate localization. \textit{Second, insufficient cross-modal interactions between video and sentence features.} Current approaches often rely on contrastive learning to enhance robustness against novel compositions by constructing negative samples, but they lack richly bidirectional interactions needed to capture fine-grained correlations between video and sentence features. Such interactions are crucial for the model to perceive variations of primitives in sentences.    

To bridge these limitations, we propose a novel Prompt-guided Primitive Alignment (PPA) framework for compositional temporal grounding. The framework includes two key components, \emph{i.e.}, the Bidirectional Masked Feature Modeling (BMFM) and the Primitive Prompt Feature Modulation (PPFM). To effectively model cross-modal interactions between video and sentence features, the BMFM is introduced for jointly reconstructing masked word tokens from visual cues and masked video tokens from textual cues via a co-attention mechanism. Such bidirectional interactions are more effective in establishing detailed correlations between modalities. To further improve compositional generalization capability, we explicitly align textual primitives with regions of target moments through the PPFM mechanism. Specifically, the module converts the similarity map between each primitive prompt and video into learnable Gaussian weights. Then, Gaussian weights are applied to emphasize the relevant regions within the video. To validate the effectiveness of the proposed PPA, we conduct extensive experiments on Charades-CG and ActivityNet-CG \cite{visa_cvpr2022} datasets, and the results demonstrate that PPA not only achieves superior compositional generalization but also generalizes well to queries containing novel words.

Overall, our main contributions are summarized as follows:
\begin{itemize}
    \item[(1)] We propose a Primitive Prompt Feature Modulation (PPFM) mechanism that explicitly aligns textual primitives with regions of target moments through prompt-guided Gaussian weighting, enabling precise moment localization under novel compositions.
    
    \item[(2)] We introduce a Bidirectional Masked Feature Modeling (BMFM) strategy that enhances cross-modal interactions by jointly reconstructing masked primitives in videos and sentences via a shared co-attention mechanism.
    \item[(3)] Extensive experiments on two benchmark datasets demonstrate that the proposed PPA achieves state-of-the-art performance, exhibiting strong generalization to queries with unseen words and compositions.
\end{itemize}

\section{Related Work}

\subsection{Video Temporal Grounding}

Video temporal grounding (VTG) is originally proposed by Gao et al. \cite{gao2017tall} and Hendricks et al. \cite{anne2017localizing}, with the objective of retrieving the correlated moment from an untrimmed video based on a given sentence query. Pioneering works adopt the two-stage pipeline, \emph{i.e.}, proposal-rank, to select the most similar candidate moment from the pre-generated proposals. For example, CTRL \cite{gao2017tall} and MCN \cite{anne2017localizing} utilized sliding windows to generate proposals in videos. TGN \cite{chen2018temporally} and SCDM \cite{yuan2019semantic} employed handcrafted predefined anchors to generate multi-scale proposals in video features. Recently, 2D-TAN \cite{2dtan_2020aaai} and MS-2D-TAN \cite{zhang2021multi} proposed to use a 2D adjacent map to enumerate all moment proposals in the video. Based on the efficient proposal representations of the 2D adjacent map, UniSDNet \cite{hu2025unified} integrates both static and dynamic modeling to learn semantic associations between videos and sentences. MMN \cite{mmn_2022aaai} further proposes to directly optimize the semantic distance between proposals and the textual query with metric learning in a common embedding space.  

Although two-stage approaches achieve high accuracy for moment localization, they suffer from redundant proposals and a severe imbalance between positive and negative samples. In contrast, one-stage approaches based on proposal-learnable or proposal-free manners have become the dominant focus in the VTG research. Inspired by the success of DETR \cite{carion2020end} that used a set of learnable query tokens to predict the locations of objects via implicit iteration on the transformer decoder, proposal-learnable methods \cite{sul2023mr,xiao2024bridging,jung2025background,yu2025hierarchical} have been widely explored based on end-to-end transformer models. To avoid learnable query tokens explicitly fit the location distribution of target moments, Momentdiff \cite{li2023momentdiff} proposes to generate the temporal boundaries of the target moment from random noises with a diffusion-based model. 
Unlike proposal methods that focus on designing complex query generation mechanisms, proposal-free methods formulate VTG as a regression problem and directly predict the timestamps of target moments. For example, ABLR \cite{yuan2019find} introduces a co-attention mechanism to model multimodal interactions between videos and sentences, and directly regresses the start and end times from attention weights or attended features. LGI \cite{lgi_cvpr2020} extracts multiple semantic phrases from textual queries and models local-global interactions to align multimodal features. VSLNet \cite{vlsnet_2020alc} predicts moment boundaries by modeling the probability distributions over all possible start and end timestamps. Different from these one-stage methods, our approach predicts the center and regresses the duration of the target moment, inspired by the object detection paradigm of CenterNet \cite{zhou2019objects}.

\subsection{Compositional generalization}

Compositional generalization refers to the ability of models to understand novel combinations of known concepts, a hallmark of human cognition and a crucial capability for robust vision-language understanding. In the context of video understanding, this ability is crucial due to the dynamic and combinatorial nature of temporal events. In the last few years, a growing number of works in related domains have demonstrated compositional generalization across several tasks, such as action recognition \cite{li2024c2c,yun2024compositional}, video question answering \cite{yu2023anetqa,liao2024align}, and video-text retrieval \cite{madasu2024icsvr}. These works are required to understand or predict unseen events by recombining known visual and linguistic components. 

Recently, to assess the compositional generalization of existing VTG methods, Li et al. \cite{visa_cvpr2022} first introduced a compositional temporal grounding task and constructed two new benchmarks, \emph{i.e.}, Charades-CG and ActivityNet-CG. To solve this problem, VISA \cite{visa_cvpr2022} decomposes the video and sentence into multiple structured hierarchies (\emph{i.e.}, global events, local actions, and atomic objects), and devises a variational cross-graph to reason fine-grained correspondence among them. SISE \cite{sesi_cvpr2023} proposes a self-supervised framework for evaluating the model's sensitivity to equivariant and invariant samples, which are generated by masking critical and irrelevant primitives of the video and text. DeCo \cite{deco_iccv2023} decomposes the textual query into sub-sentences with different granularities and learns to align them with the video using a mask-and-predict contrastive loss. SHINE \cite{shine_ceev2024} proposes a saliency-aware hierarchical negative ranking strategy that incorporates both semantically similar and dissimilar negative samples to enhance compositional generalization in VTG. VID \cite{vdi_cvpr2023} and CFRM \cite{cfrm_2024wacv} leverage pretrained V\&L models to enhance video-sentence alignment, thereby achieving significant improvements in handling unseen words in sentences. Unlike these methods, our approach introduces a novel framework to enhances compositional generalization by establishing fine-grained correspondences between textual primitives and target moments. 

\begin{figure*}[t]
\centering
\includegraphics[width=\linewidth]{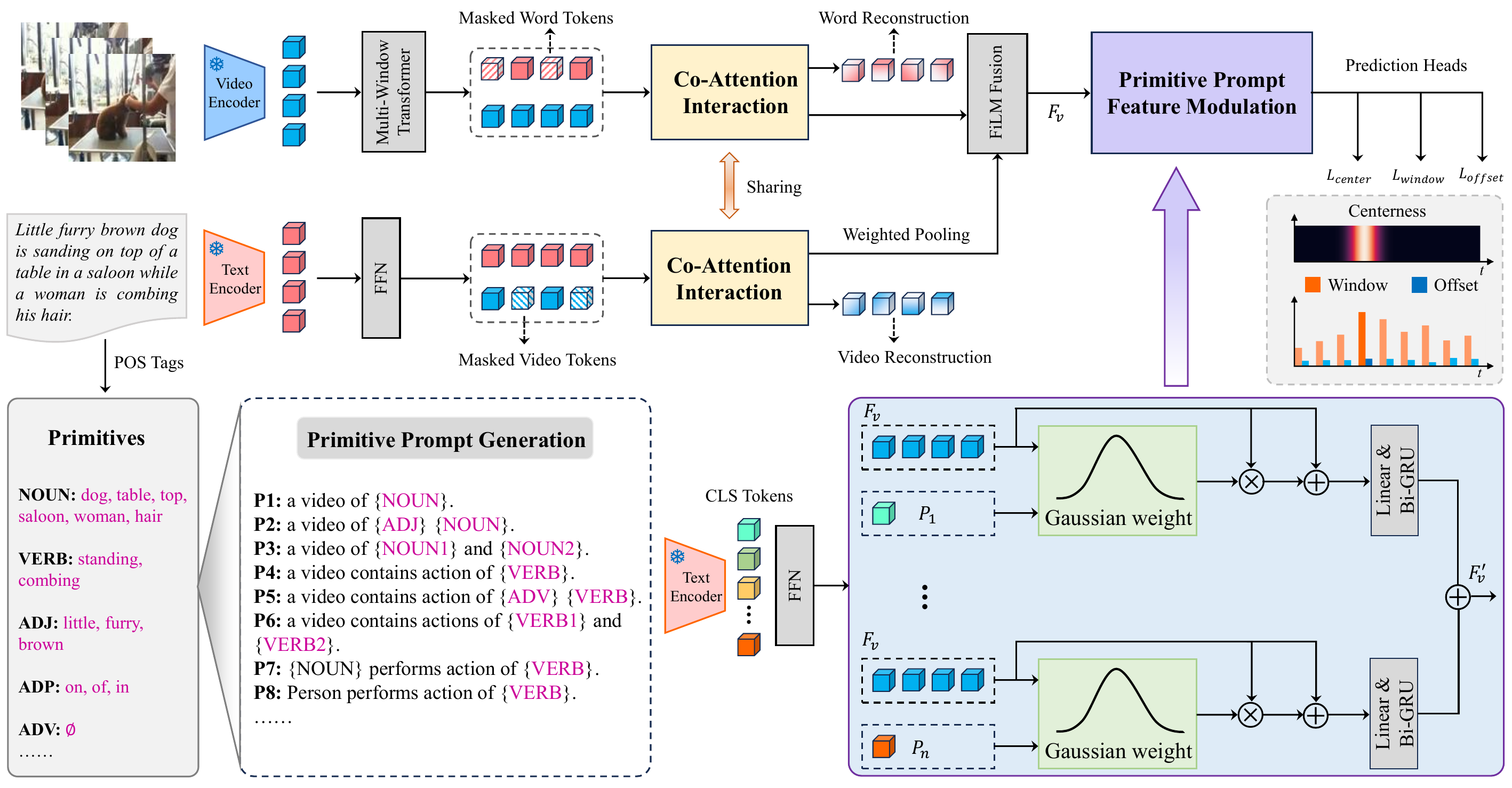}
\caption{The overall network architecture of the proposed  Prompt-guided Primitive Alignment (PPA) framework.}
\label{fig:figure2}
\vspace{-0.2cm}
\end{figure*}

\section{Methodology}
\label{methodology}

In this section, we first provide the basic definition of compositional temporal grounding and present an overview of the proposed framework. Next, we elaborate on the details of the primary components in the framework. Finally, we conduct extensive experiments and ablation studies on available datasets. 

\subsection{Problem Definition and Overview}
\label{overview}

The goal of compositional temporal grounding is to localize the temporal boundary $(t_{s}, t_{e})$ of the video moment most relevant to a given query, where the query in the test set contains novel compositions or novel words unseen during training. Here, $t_{s}$ and $t_{e}$ denote the start and end times of the target moment, respectively. The input video is represented as $\mathcal{V}=\{v_{i}\}_{i=1}^{T} \in \mathbb{R}^{T \times{D}_{v}}$, where $T$ is the total number of video clips and $D_{v}$ is the dimensionality of the visual features. The sentence query is represented as $\mathcal{Q}=\{q_{i}\}_{i=1}^{M} \in \mathbb{R}^{M \times D_{q}}$, where $M$ is the length of the query and $D_{q}$ is the dimensionality of the linguistic features. 

As illustrated in Figure \ref{fig:figure2}, the overall architecture of the proposed  Prompt-guided Primitive Alignment (PPA) framework consists of three components: (1) video and language feature encoding, (2) bidirectional masking feature modeling, and (3) primitive prompt feature enhancement. In the following sections, we describe each component in detail, together with the training and inference procedures.

\subsection{Video and Sentence Feature Encoding}
\label{encoding}

For video feature encoding, we first employ a pre-trained video feature encoder \cite{c3d_cvpr2015,i3d_cvpr2017} to extract features from divided video clips. The extracted features are further projected through linear layers and denoted as $\mathbf{F}_{v} \in \mathbb{R}^{T\times D}$. Since the importance of modeling temporal context for action localization \cite{non-local_cvpr2018,bahrami2023much}, we design a Multi-Window Transformer (MWFormer) to encode $\mathbf{{F}}_{v}$. The MWFormer builds on temporal windowed attention \cite{bahrami2023much}, which jointly models long-range and local dependencies. Specifically, it employs $N$ temporal windows of varying sizes $\{W_{i}\}_{i=1}^{N}$. For each window, we partition the feature sequence into non-overlapping segments with length of $W_{i}$. Instead of computing the attention over the entire sequence of length $T$, we compute the attention $\frac{T}{W_{i}}$ times, where each query $\mathbf{Q} \in \mathbb{R}^{W \times D}$ corresponds to each segment. The keys and values are two adjacent segments, \emph{i.e.}, $\mathbf{K}, \mathbf{V} \in \mathbb{R}^{2W_{i} \times D}$. Therefore, the calculation of attention in each segment is formulated as:  
\begin{equation}
    \mathcal{F}(\mathbf{Q},\mathbf{K},\mathbf{V})=Softmax(\frac{\mathbf{QK}^{\top}}{\sqrt{D}})\mathbf{V}.
\end{equation}

Here, the attention for the different windows $\{W_{i}\}_{i=1}^{N}$ is computed in a parameter-sharing manner, which produces a set of weighted video features $\{\mathbf{F}_{v}^{i}\}_{i=1}^{N}$. These features are then summed and passed through a feed-forward network (FFN). Formally, this can be expressed as: 
\begin{equation}
    \mathbf{F_{v}}=FFN(\sum_{i=1}^{N}\mathbf{F}_{v}^{i}).  
\end{equation}

Compared with vanilla transformer blocks with self-attention, the MWFormer block not only captures multi-scale local context but also reduces computational complexity for long video feature sequences. In our model, the number of windows is set to $N=3$, and the number of MWFormer layers is set to 2.

For linguistic feature encoding, we employ a pre-trained language encoder to extract word-level features from the input sentence queries. Specifically, to mitigate the semantic gap between video and text and to enhance the representation of novel words in test queries, we adopt the CLIP \cite{radford2021learning} text encoder as our sentence query encoder. The extracted word features are further projected through linear layers and denoted as $\mathbf{F}_{q} \in \mathbb{R}^{M \times D}$. 

\subsection{Bidirectional Masked Feature Modeling}
\label{masking}

In video moment localization, the semantics of the sentence query are typically relevant to only a portion of the video content. This means that videos often contain richer and more diverse semantic elements than their accompanying textual descriptions. To achieve more accurate grounding results, it is essential for the model to perform fine-grained cross-modal interactions, which thoroughly understand the fundamental elements of both modalities and establish precise semantic alignment between them. To this end, we propose a bidirectional masked feature modeling strategy that enhances the representation capability of both video and sentence features. After that, the sentence features are dynamically mapped into the video domain, yielding semantically enriched multi-modal video representations.

Specifically, we first introduce a co-attention module to model the mutual interactions between video and sentence features. Formally, the interactions are computed as:
\begin{equation}
    \mathbf{F}_{v}^{(Q)}=\mathbf{W}^{(v,Q)} \mathbf{F}_{v}, \quad \mathbf{F}_{w}^{(Q)}=\mathbf{W}^{(w,Q)} \mathbf{F}_{w},
\end{equation}
\begin{equation}
    \mathbf{F}_{v}^{(V)}=\mathbf{W}^{(v,V)} \mathbf{F}_{v}, \quad
    \mathbf{F}_{w}^{(V)}=\mathbf{W}^{(w,V)} \mathbf{F}_{w},
\end{equation}
\begin{equation}
    \mathbf{A} =\frac{\mathbf{F}_{v}^{(Q)}\mathbf{F}_{w}^{(Q)}}{\sqrt{D}},
\end{equation}
\begin{equation}
    \mathbf{F}_{v2w}=Softmax(\mathbf{A}^{\top})\mathbf{F}_{v}^{(V)}, \quad \mathbf{F}_{w2v}=Softmax(\mathbf{A})\mathbf{F}_{w}^{(V)},
\end{equation}
\begin{equation}
    \mathbf{F}_{v}^{'}=\mathbf{F}_{v}+\mathbf{W}^{(v)}\mathbf{F}_{w2v}, \quad \mathbf{F}_{w}^{'}=\mathbf{F}_{t}+\mathbf{W}^{(w)}\mathbf{F}_{v2w},
\end{equation}
where $\mathbf{W}^{(v,Q)}$, $\mathbf{W}^{(w,Q)}$, $\mathbf{W}^{(v,V)}$, $\mathbf{W}^{(t,V)}$, $\mathbf{W}^{(v)}$ and $\mathbf{W}^{(t)}$ are learnable parameters. The $V$ and $K$ indicate value and key, respectively. $A \in \mathbb{R}^{T \times M}$ denotes the similarity matrix between video and sentence features.

To realize bidirectional masked feature modeling, we employ two co-attention modules with shared weights. Given the video features $\mathbf{F}_{v} \in \mathbb{R}^{T \times D}$ and sentence features $\mathbf{F}_{w} \in \mathbb{R}^{M \times D}$, the first co-attention module operates by randomly masking a subset of word features, where the masked tokens are replaced with a learnable token $[\mathbf{MASK}]$. The video features and the masked textual features are first processed by the feature encoding module introduced in the previous section, and then fed into the co-attention module. This module leverages visual cues from the video to reconstruct the masked word features. Similarly, in the second co-attention module, we randomly mask a subset of the video features that fall within the region described by the text. The masked video features and the complete textual features are fed into the co-attention module, which reconstructs the masked video frame features by attending to the semantic cues in the sentence.

For the masked video frame and word features, we adopt a cosine similarity loss to guide their reconstruction. During loss computation, the original features extracted from pre-trained encoders serve as reconstruction targets. Let $\mathbf{F}_{v}^{(i,\text{rec})}, \ \mathbf{F}_{v}^{(i,\text{raw})} \in \mathbb{R}^{D_v}$ denote the reconstructed and original features of the $i$-th masked video frame, respectively. Here $i$ indicates the masked positions and $M_v$ is the number of masked video frames. The reconstruction loss on the masked video frames is defined as:
\begin{equation}
    \mathcal{L}_{v} = 1 - \frac{1}{B}\sum_{j=1}^{B}(\frac{1}{M_{v}}\sum_{i=1}^{M_{v}} \frac{\mathbf{F}_{v}^{(i,rec)} \cdot sg(\mathbf{F}_{v}^{(i, raw)})}{\|\mathbf{F}_{v}^{(i,rec)}\| \cdot \|\mathbf{F}_{v}^{(i, raw)}\|}),
\end{equation}

Similarly, let $\mathbf{F}_{t}^{(i,\text{rec})}, \ \mathbf{F}_{t}^{(i,\text{raw})} \in \mathbb{R}^{D_t}$ denote the reconstructed and original features of the $i$-th masked word, where $i$ indicates the masked positions and $M_t$ is the number of masked words. The reconstruction loss on the masked words is defined as:
\begin{equation}
    \mathcal{L}_{t} = 1 - \frac{1}{B}\sum_{j=1}^{B}(\frac{1}{M_{t}}\sum_{i=1}^{M_{t}} \frac{\mathbf{F}_{t}^{(i,rec)} \cdot sg(\mathbf{F}_{t}^{(i, raw)})}{\|\mathbf{F}_{t}^{(i,rec)}\| \cdot \|\mathbf{F}_{t}^{(i, raw)}\|}),
\end{equation}

Thus, the overall feature reconstruction loss for bidirectional masked feature modeling is defined as:
\begin{equation}
    \mathcal{L}_{rec} = \mathcal{L}_{v} + \mathcal{L}_{t},
\end{equation}
where $B$ is the batch size, and $sg(\cdot)$ denotes the stop-gradient operation.

After the bidirectional interactions between video and sentence features, we fuse the enhanced video and sentence features using a dynamic mechanism. Specifically, we first apply a self-attention pooling operation to aggregate the word-level features $\mathbf{F}_{w} \in \mathbb{R}^{M\times D}$ into a global representation $\mathbf{F}_{w}^{g} \in \mathbb{R}^{D}$. Subsequently, $\mathbf{F}_{w}^{g}$ is projected through two linear layers to generate two modulation vectors, denoted as $\bm{\mu} \in \mathbb{R}^{D}$ and $\bm{\nu} \in \mathbb{R}^{D}$. Formally, the dynamic fusion of video and sentence features is defined as:
\begin{equation}
    \bm{\mu} = tanh(\mathbf{W}_{\mu} \mathbf{F}_{t}^{g} + \mathbf{b}_{\mu}), \quad
    \bm{\nu} = tanh(\mathbf{W}_{\nu} \mathbf{F}_{t}^{g} + \mathbf{b}_{\nu}),
\end{equation}
\begin{equation}
     {\mathbf{\hat{F}}_{v}} = ReLU(\bm{\mu} \odot \mathbf{F}_{v}^{'} + \bm{\nu}),
\end{equation}
where $\mathbf{W}_{\mu}$, $\mathbf{b}_{\mu}$, $\mathbf{W}_{\nu}$, and $\mathbf{b}_{\nu}$ are learnable parameters. $\odot$ means the element-wise multiplication.

\subsection{Primitive Prompt Feature Modulation}
\label{sec:prompt}

The ability to capture variations in textual primitives and their compositions is crucial for achieving compositional generalization. To this end, we propose explicitly modeling the semantic correlation between textual primitives and their corresponding video moments. This correlation is then used to reweight the fused multi-modal video features $\mathbf{\hat{F}}_{v}$, guiding the model to attend more effectively to semantically aligned content.

Specifically, given an input sentence, for example, ``\textit{Little furry brown dog is standing on top of a table in a saloon while a woman is combing his hair}''. Our method first extracts words belonging to different parts of speech using SpaCy and treats them as textual primitives. We then define a set of prompt templates manually, such as “a video of $\{\mathrm{NOUN}\}$”, “a video contains action of $\{\mathrm{VERB}\}$”, or “a video of $\{\mathrm{ADJ} \ {\mathrm{NOUN}}\}$”, as shown in Figure~\ref{fig:figure2}. By randomly filling these templates with extracted primitives, we generate primitive prompts $P = \{P_1, P_2, ..., P_n\}$, primarily based on nouns, verbs, adjectives, and adverbs that represent entities or attributes.  

After constructing the primitive prompts, we leverage the image-text pre-trained model CLIP~\cite{radford2021learning} to extract their feature representations. Specifically, the output corresponding to the [CLS] token is adopted as the representation of each prompt. The obtained primitive prompt features are subsequently fed into a  multilayer perceptron (MLP), resulting in a set of vectors $\mathbf{P} = \{\mathbf{p}_{1}, \mathbf{p}_{2}, \ldots, \mathbf{p}_{n}\}$, where $\mathbf{p}_{i} \in \mathbb{R}^{D}$ denotes the feature of the $i$-th prompt, and $n$ is the total number of generated primitive prompts.

Since these textual primitives are semantically relevant to the target video moment, their features can be exploited to highlight the corresponding region within the video. For each primitive prompt $\mathbf{p}_{i}$, we project both $\mathbf{p}_{i}$ and the fused multi-modal video feature $\mathbf{\hat{F}}_{v}$ into a shared semantic space using a linear layer transformation followed by L2 normalization. For clarity, we denote the mapped features as $\mathbf{p}_{i} \in \mathbb{R}^{1 \times D}$ and $\mathbf{\hat{F}}_{v} \in \mathbb{R}^{T \times D}$, where $T$ is the length of video features. The semantic similarity between $\mathbf{p}_{i}$ and $\mathbf{\hat{F}}_{v}$ is then measured via cosine similarity, and subsequently transformed into Gaussian weights through a learnable Gaussian function. Formally, this process is defined as:
\begin{equation}
    w_{i} = \tau \cdot exp(- \frac{(1-\mathbf{p}_{i} \mathbf{\hat{F}}_{v})^{\top}}{2\rho^{2}}),
\end{equation}
where $\tau$ and $\rho$ are learnable parameters, and $w_{i} \in \mathbb{R}^{T \times 1}$ denotes the Gaussian weight reflecting the semantic relevance between the $i$-th prompt and the multi-modal features. Subsequently, we modulate the fused multi-modal features based on these weights, thereby highlighting regions that correspond to the target video moment. The enhanced feature guided by the $i$-th primitive prompt is given by: 
\begin{equation}
    \mathbf{\hat{F}}_{v}^{i} = \mathbf{\hat{F}}_{v} + {w}_{i}\mathbf{\hat{F}}_{v},
\end{equation}
where $\mathbf{\hat{F}}_{v}^{i}$ is the prompt-enhanced multi-modal feature corresponding to the $i$-th primitive. By repeating this process for all primitive prompts, we obtain a set of modulated representations, \emph{i.e.}, $\{\mathbf{\hat{F}}_{v}^{1}, \mathbf{\hat{F}}_{v}^{2}, ..., \mathbf{\hat{F}}_{v}^{n}\}$.

To capture temporal dependencies on prompt-modulated video features, we apply a bidirectional Gated Recurrent Unit (Bi-GRU) with shared weights on $\{\mathbf{\hat{F}}_{v}^{1}, \mathbf{\hat{F}}_{v}^{2}, ..., \mathbf{\hat{F}}_{v}^{n}\}$. The final fused representation for target moment prediction is obtained by summing the Bi-GRU output:
$\mathbf{\hat{F}}_{v}^{o}=\mathbf{\hat{F}}_{v}^{1}+\mathbf{\hat{F}}_{v}^{2}+...+\mathbf{\hat{F}}_{v}^{n}$.

Furthermore, to ensure that each primitive prompt attends precisely to the relevant video region, we supervise the learnable Gaussian weights via Kullback–Leibler (KL) divergence. Specifically, we construct a 1D Gaussian distribution centered at the ground-truth moment as the target distribution $\hat{w}$, and define the loss as:
\begin{equation}
    \mathcal{L}_{h} = \sum_{j=1}^{n} D_{KL}(\hat{w}\|w)=\sum_{j=1}^{n}\sum_{i=1}^{T}\hat{w}_{i}log(\frac{\hat{w}_{i}}{w_{i}}),
\end{equation}
where $T$ is the number of video clips, $\hat{w}_i$ is the target Gaussian weight at position $i$, and $w_i$ is the predicted weight.

\subsection{Training and Inference}

Following previous work \cite{liu2022umt}, our method localizes the target video moment by directly predicting its center and duration. To predict the center, we first quantize the ground-truth center index $c$ into $\Tilde{c}$, and construct a one-dimensional Gaussian kernel $H_{x} = \exp\left(-\frac{(x - \Tilde{c})^2}{2\sigma_c^2}\right)$, where $x$ denotes the temporal coordinate and $\sigma_c$ is a standard deviation relevant to the ground-truth duration $d$. This yields a heatmap $H \in [0,1]^{T}$, where $T$ is the number of temporal clips. The center prediction is supervised using the Gaussian Focal Loss~\cite{zhou2019objects}, which emphasizes learning on central positions while suppressing irrelevant locations: 
\begin{equation}
    \mathcal{L}_{c} = - \sum_{x=0}^{L_v}\left\{
    \begin{array}{ll}
     (1-\hat{H}_{x})^{\alpha}log(\hat{H}_{x}), \quad \quad \quad \quad \quad  If \ H_{x}=1 \\
     (1-\hat{H_{x}})^{\beta} \hat{H}_{x}^{\alpha}log(1-\hat{H}_{x}),   \quad \quad Otherwise 
     
    \end{array}
    \right.
\end{equation}
where $\alpha$ and $\beta$ denote the weighting and focusing parameters of the focal loss, respectively set to 2.0 and 4.0. For duration $d$ and offset $o$, we adopt the Smooth L1 loss for regression. The overall loss for the target moment prediction is defined as the sum of the three components:
\begin{equation}
    \mathcal{L}_p = \mathcal{L}_{c} + \lambda_{d}\mathcal{L}_{d} + \lambda_{o}\mathcal{L}_{o},
\end{equation}
where $\lambda_{d}$ and $\lambda_{o}$ are weighting coefficients for the duration and offset loss terms, respectively. In our experiments, we set $\lambda_{d} = 0.1$ and $\lambda_{o} = 1.0$.

During training, the overall optimization objective of the model is defined as:
\begin{equation}
    \mathcal{L} = \mathcal{L}_{p} + \lambda_{rec} \mathcal{L}_{rec} + \lambda_{h} \mathcal{L}_{h},
\end{equation}
where $\lambda_{rec}$ and $\lambda_{h}$ are balance parameters for the reconstruction loss and KL loss, respectively. In our experiments, we set $\lambda_{rec}=2.0$ and $\lambda_{h}=5.0$.

During inference, the start and end boundaries of the target video moment are derived from the predicted center $c$, duration $d$, and offset $o$:
\begin{equation}
    (t_{s}, t_{e}) = \frac{S}{T}(c+o-\frac{d}{2}, \ c+o+\frac{d}{2}),
\end{equation}
where $S$ denotes the entire video duration in seconds, and $T$ is the length of the video feature sequence.

\section{Experiments}
\label{experiments}

\subsection{Datasets and Evaluation}
\textbf{Datasets}. 
We conduct experiments on two benchmark datasets for compositional temporal grounding, \emph{i.e.}, Charades-CG and ActivityNet-CG \cite{visa_cvpr2022}. Both datasets are re-constructed from the widely used temporal grounding benchmarks, Charades-STA \cite{gao2017tall} and ActivityNet-Captions \cite{krishna2017dense}, respectively. Each dataset is divided into a training set and three test sets. Specifically, the \emph{Test-Trivial} and \emph{Novel-Word} sets are designed to evaluate generalization on seen and unseen words, while the \emph{Novel-Composition} set is used to assess compositional generalization. Charades-CG contains 8,281 / 3,096 / 703 / 3,442 video–sentence pairs for the \emph{training} / \emph{Test-Trivial} / \emph{Novel-Word} / \emph{Novel-Composition} splits. ActivityNet-CG consists of 36,724 / 15,712 / 3,944 / 12,028 video–sentence pairs for \emph{training} / \emph{Test-Trivial} / \emph{Novel-Word} / \emph{Novel-Composition} split.

\textbf{Evaluation}. 
Following previous work \cite{gao2017tall, visa_cvpr2022}, we adopt $Rank@1, IoU=m$ and $mIoU$ as evaluation metrics. Specifically, $Rank@1, IoU=m$ measures the percentage of samples for which at least one top-1 localized result has an IoU with the ground truth greater than $m$, where $m \in \{0.5, 0.7\}$. In contrast, $mIoU$ denotes the average IoU over all samples.

\subsection{Implementation Details}

In our work, we adopt pre-trained video and language encoders to extract video and sentence features. For video feature extraction, we follow prior work \cite{lgi_cvpr2020} and utilize the pre-trained I3D model \cite{i3d_cvpr2017} for the Charades-CG dataset, while for the ActivityNet-CG dataset, we follow \cite{vdi_cvpr2023,cfrm_2024wacv} and use the visual encoder of CLIP (ViT-B/32) \cite{radford2021learning}. The dimensions of the extracted video features are 1024 for Charades-CG and 512 for ActivityNet-CG. To ensure consistency, feature sequences longer than 200 are uniformly downsampled to a length of 200. For sentence feature extraction, we employ the CLIP (ViT-B/32) language encoder to obtain word-level features from the sentence and prompt features from primitive prompts, both with dimensionality 512. Our method is implemented in PyTorch and optimized using the Adam optimizer \cite{kingma2014adam}. For Charades-CG, we set the batch size to 32 and the learning rate to $5.5 \times 10^{-4}$. For ActivityNet-CG, the batch size is 64 and the learning rate is $6.0 \times 10^{-4}$.

\subsection{Comparision with the State-of-the-Arts}

To evaluate the effectiveness of our proposed method, we compare the proposed PPA with state-of-the-art approaches on the Charades-CG and ActivityNet-CG datasets. These methods can be grouped into three categories: (1) Reinforcement learning (RL)-based methods: TSP-PRL \cite{tsp_aaai2020}; (2) Proposal-based (PB) methods: TMN \cite{tmn_2018eccv}, 2D-TAN \cite{2dtan_2020aaai}, MMN \cite{mmn_2022aaai}, VDI \cite{vdi_cvpr2023}, SESI (MS-2D-TAN) \cite{sesi_cvpr2023}, and CFRM \cite{cfrm_2024wacv}; (3) Proposal-free (PF) methods: LGI \cite{lgi_cvpr2020}, VLSNet \cite{vlsnet_2020alc}, VISA \cite{visa_cvpr2022}, DeCo \cite{deco_iccv2023}, SHINE (Moment-DETR) \cite{shine_ceev2024}, and MESM \cite{mesm_aaai2024}.

\begin{table}[t]
\begin{center}
\tiny

  \caption{Comparison with state-of-the-art methods on Charades-CG dataset.}
  \label{tab:charades-cg}
  \tiny
  \setlength{\tabcolsep}{1.2mm}{
  \begin{tabular}{llccccccccccc}
    \toprule
    
    \multirow{2}{*}{}  &\multirow{2}{*}{\textbf{Methods}} &\multirow{2}{*}{\textbf{Years}}  &\multicolumn{3}{c}{\textbf{Test-Trivial}}  &\multicolumn{3}{c}{\textbf{Novel-Composition}}  &\multicolumn{3}{c}{\textbf{Novel-Word}} \\   
    \cmidrule[0.6pt](lr){4-6}
    \cmidrule[0.6pt](lr){7-9}
    \cmidrule[0.6pt](lr){10-12}
    & &  & R1@0.5  & R1@0.7 & mIoU   & R1@0.5  & R1@0.7 & mIoU   & R1@0.5  & R1@0.7  & mIoU \\
   
    \midrule
    RL        & TSP-PRL\cite{tsp_aaai2020}         &2020    & 39.86    & 21.07   & 38.41  & 16.30    & 2.04    & 13.52  & 14.83    & 2.61     & 14.03 \\
    \midrule
    \multirow{6}{*}{PB}  & TMN \cite{tmn_2018eccv} &2018    & 18.75    & 8.16    & 19.82  & 8.68     & 4.07    & 10.14  & 9.43     & 4.96     & 11.23 \\
     & 2D-TAN \cite{2dtan_2020aaai}                &2020    & 48.58    & 26.49   & 44.27  & 30.91    & 12.23   & 29.75  & 29.36    & 13.21    & 28.47 \\
     & MMN \cite{mmn_2022aaai}                     &2022    & -        & -       & -      & -        & -       & -      & 43.85    & 24.17    & 39.50 \\
     & VDI \cite{vdi_cvpr2023}                     &2023    & -        & -       & -      & -        & -       & -      & 46.47    & 28.63    & 41.60 \\
     & SESI \cite{sesi_cvpr2023}                   &2023    & 58.14    & 37.98   & 50.58  & 46.54    & 25.10   & 40.00  & 50.36    &28.78     & 43.15 \\
     & CFRM \cite{cfrm_2024wacv}                   &2024    & -        & -       & -      & -        & -       & -      & 45.04    & 21.44    & -     \\
     
    \midrule
    \multirow{7}{*}{PF}  & LGI \cite{lgi_cvpr2020} &2020    & 49.45    & 23.80   & 45.01  & 29.42    & 12.73   & 30.09  & 26.48    & 12.47    & 27.62 \\ 
    & VLSNet \cite{vlsnet_2020alc}                 &2020    & 45.91    & 19.80   & 41.63  & 24.25    & 11.54   & 31.43  & 25.60    & 10.07    & 30.21 \\
    & VISA \cite{visa_cvpr2022}                    &2022    & 53.20    & 26.52   & 47.11  & 45.41    & 22.71   & 42.03  & 42.35    & 20.88    & 40.18 \\ 
    & DeCo \cite{deco_iccv2023}                    &2023    & 58.75    & 28.71   & 49.06  & 47.39    & 21.06   & 40.70  & -        & -        & -     \\
    & SHINE \cite{shine_ceev2024}                  &2024   & 57.14     & 33.85   & 49.32 
     & 44.65    & 23.21   & 39.86  & 47.05    & 24.32    & 41.57 \\
    & MESM \cite{mesm_aaai2024}                    &2024    & -        & -       & -      & 44.39    & 23.27   &
    39.89  & \textbf{52.66}    & 31.22    & 46.38  \\

    & Ours                                         &2025    & \textbf{62.31}    & \textbf{42.99}   & \textbf{53.68}  & \textbf{49.13}    & \textbf{29.17}   & \textbf{43.53}  & 51.22    & \textbf{33.67}    & \textbf{46.44} \\
    \bottomrule
  \end{tabular} }

\end{center}
\end{table}

\textbf{Comparisons on Charades-CG.} 
As shown in Table \ref{tab:charades-cg}, we compare the proposed PPA method with existing approaches across three evaluation splits of the Charades-CG dataset. It can be observed that: (1) On the Test-Trivial split, PPA significantly outperforms existing methods across all metrics. In particular, compared with the best-performing method DeCo, PPA achieves improvements of 3.56\%, 14.28\%, and 4.62\% in $R1@0.5$, $R1@0.7$, and $mIoU$, respectively. (2) On the Novel-Composition split, our method achieves superior performance compared with those approaches \cite{sesi_cvpr2023,visa_cvpr2022,deco_iccv2023,shine_ceev2024,mesm_aaai2024} that are specifically designed for compositional generalization of video temporal grounding. Notably, compared with VISA, PPA yields improvements of 3.72\%, 6.46\%, and 1.5\% on $R1@0.5$, $R1@0.7$, and $mIoU$, respectively, highlighting its strong generalization ability while preserving high localization accuracy. (3) On the Novel-Word split, PPA achieves competitive results against the state-of-the-art MESM, achieving improvements of 2.45\% and 0.06\% in $R1@0.7$ and $mIoU$, respectively. 

\begin{table}[h]
\begin{center}
\tiny
  \caption{Comparison with state-of-the-art methods on ActivityNet-CG dataset.}
  \label{tab:anet-cg}
  \setlength{\tabcolsep}{1.0mm}{
  \begin{tabular}{llccccccccccc}
    \toprule
    
    \multirow{2}{*}{}  &\multirow{2}{*}{\textbf{Methods}} &\multirow{2}{*}{\textbf{Years}}  &\multicolumn{3}{c}{\textbf{Test-Trivial}}  &\multicolumn{3}{c}{\textbf{Novel-Composition}}  &\multicolumn{3}{c}{\textbf{Novel-Word}} \\   
    \cmidrule[0.6pt](lr){4-6}
    \cmidrule[0.6pt](lr){7-9}
    \cmidrule[0.6pt](lr){10-12}
    & & & R1@0.5  & R1@0.7 & mIoU   & R1@0.5  & R1@0.7 & mIoU   & R1@0.5  & R1@0.7  & mIoU \\
   
    \midrule
    RL                  & TSP-PRL \cite{tsp_aaai2020}  &2020  & 34.27  & 18.80  & 37.05  & 14.74  & 1.43  & 12.61  & 18.05  & 3.15  & 14.34 \\
    \midrule
    \multirow{5}{*}{PB} & TMN \cite{tmn_2018eccv}      &2018  & 16.82  & 7.01   & 17.13  & 8.74   & 4.39  & 10.08  & 9.93   & 5.12  & 11.38 \\
                    & 2D-TAN \cite{2dtan_2020aaai}     &2020  & 44.50  & 26.03  & 42.12  & 22.80  & 9.95  & 28.49  & 23.86  & 10.37 & 28.88 \\
                    & MMN \cite{mmn_2022aaai}          &2022  & -      & -      & -      & -      & -     & -      & 31.05  & 15.48 & 33.16 \\
                    & VDI \cite{vdi_cvpr2023}          &2023  & -      & -      & -      & -      & -     & -      & 32.35  & 16.02 & 34.32 \\
                    & CFRM \cite{cfrm_2024wacv}        &2024  & -      & -      & -      & -      & -     & -      & 24.57  & 10.54 & -     \\
    \midrule
    \multirow{7}{*}{PF}     & LGI \cite{lgi_cvpr2020}  &2020     & 43.56  & 23.29  & 41.37  & 23.21  & 9.02  & 27.86  & 23.10  & 9.03  & 26.95 \\ 
                    & VLSNet \cite{vlsnet_2020alc}     &2020  & 39.27  & 23.12  & 42.51  & 20.21  & 9.18  & 29.07  & 21.68  & 9.94  & 29.58 \\
                    & VISA \cite{visa_cvpr2022}        &2022  & 47.13  & 29.64  & 44.02  & 31.51  & 16.73 & \textbf{35.85}  & 30.14  & 15.90 & \textbf{35.13} \\
                    & DeCo \cite{deco_iccv2023}        &2023  & 43.98  & 24.25  & 43.47  & 27.35  & 11.66 & 31.27  & -      & -     & -     \\
                    & DeCo+LGI \cite{deco_iccv2023}    &2023  & 47.38  & 28.43  & 46.03  & 28.69  & 12.98 & 32.67  & -      & -     & -     \\
                    & SHINE \cite{shine_ceev2024}      &2024   & 44.19 & 25.81  & 43.49 & 30.60  & 14.40 & 33.13 & 29.59   & 15.10 & 32.43 \\

                    & Ours                             &2025  & \textbf{48.76}   & \textbf{31.40}    & \textbf{46.87}     & \textbf{33.04}    & \textbf{17.23}   & 34.61    & \textbf{32.35}     & \textbf{16.45}     & 34.05   \\
    \bottomrule[1.0pt]
  \end{tabular}}
\end{center}
\end{table}

\textbf{Comparisons on ActivityNet-CG.} In Table \ref{tab:anet-cg}, we compare the proposed PPA method with existing approaches across three evaluation splits of the ActivityNet-CG dataset. The results show that: (1) On the Test-Trivial split, the proposed PPA outperforms the best-performing method VISA by 1.63\%, 1.76\%, and 2.85\% in $R1@0.5$, $R1@0.7$, and $mIoU$, respectively. (2) On the Novel-Composition split, our method achieves gains of 2.85\% and 1.53\% over VISA in $R1@0.5$, $R1@0.7$, while falling behind VISA by 1.24\% in terms of $mIoU$. (3) On the Novel-Word split, PPA falls behind VISA by 1.08\% in $mIoU$, but surpasses VISA by 2.21\% and 0.55\% in $R1@0.7$ and $mIoU$. These results demonstrate that the proposed PPA not only maintains robust performance on samples with seen words and compositions, but also generalizes effectively to samples with unseen words and novel compositions. Compared with Charades-CG, the gains on ActivityNet-CG are less pronounced, which can be attributed to the higher complexity of ActivityNet-CG, where longer queries and more diverse video scenes impose greater demands on the model’s localization capability.

\subsection{Ablation Study}
To comprehensively evaluate the effectiveness of the proposed PPA framework, we conduct extensive ablation studies on both the Charades-CG and ActivityNet-CG datasets, with experimental results reported on the Novel-Composition split.

\subsubsection{Ablation Study of Each Component}

\begin{table}[h]
\begin{center}
\small
  \caption{Ablation studies of model components on the Novel Composition split.}
  \label{tab:components}
  \setlength{\tabcolsep}{1.6mm}{
  \begin{tabular}{ccccccc}
    \toprule
     \multirow{2}{*}{\textbf{MWFormer}} & \multirow{2}{*}{\textbf{BMFM}}  & \multirow{2}{*}{\textbf{PPFM}}   & \multicolumn{2}{c}{\textbf{Charades-CG}} & \multicolumn{2}{c}{\textbf{ActivityNet-CG}}   \\
     \cmidrule[0.6pt](lr){4-5}
     \cmidrule[0.6pt](lr){6-7}
     &   &  & R1@0.5   & R1@0.7   & R1@0.5  & R1@0.7 \\
     
     \hline
        \ding{56}   & \ding{56}     & \ding{56}   & 44.89     & 25.13      & 28.10      & 13.01  \\
        \ding{52}   & \ding{56}     & \ding{56}   & 44.89     & 26.12      & 30.09      & 14.26  \\
        \ding{52}   & \ding{52}     & \ding{56}   & 45.90      & 26.67     & 31.83      & 15.40  \\
        \ding{52}   & \ding{56}     & \ding{52}   & 46.34     & 27.05      & 30.64      & 14.92  \\
        \ding{56}   & \ding{52}     & \ding{52}   & 47.53     & 27.37      & 31.91      & 16.09  \\
        \ding{52}   & \ding{52}     & \ding{52}   & \textbf{49.13}     & \textbf{29.17}      & \textbf{33.04}     & \textbf{17.23}  \\

    \bottomrule
  \end{tabular}}
  \end{center}
\end{table}

In this ablation study, we validate the contribution of each component in the proposed PPA framework, including WFormer, BMFM, and PPFM. As shown in line 1 of Table \ref{tab:components}, the baseline model without these components achieves the lowest localization performance. By progressively incorporating MWFormer, BMFM, and PPFM into the baseline model, we can observe that the performance consistently improves. Specifically, adding MWFormer into the baseline yields gains of 1.99\% and 1.25\% in $R1@0.5$ and $R1@0.7$ on ActivityNet-CG, demonstrating the importance of modeling multi-scale context for accurate moment localization. Further integrating BMFM leads to an additional 1.74\% and 1.14\% improvements in $R1@0.5$ and $R1@0.7$, respectively. This demonstrates that bidirectional masked feature modeling can effectively enhance video and sentence representations by capturing primitive changes. Finally, incorporating PPFM brings further gains of 1.21\% and 1.83\% in $R1@0.5$ and $R1@0.7$ compared to the model in line 3 of Table~\ref{tab:components} on ActivityNet-CG. These results show the effectiveness of primitive prompt feature modulation in capturing fine-grained variations in textual primitives and their compositions, thereby strengthening the compositional generalization ability of video temporal grounding.

\subsubsection{Impact of Window Sizes in MWFormer}

\begin{table}[h]
\begin{center}
\small
  \caption{Impact of window sizes in MWFormer on Novel Composition split.}
  \label{tab:window_size}
  \setlength{\tabcolsep}{1.8mm}{
  \begin{tabular}{ccccccc}
    \toprule
     \multirow{2}{*}{\makecell{\textbf{Window} \\ \textbf{Sizes}}}  & \multicolumn{3}{c}{\textbf{Charades-CG}}  & \multicolumn{3}{c}{\textbf{ActivityNet-CG}}   \\
      \cmidrule[0.6pt](lr){2-4}
      \cmidrule[0.6pt](lr){5-7}
       & R1@0.5   & R1@0.7  & mIoU  & R1@0.5  & R1@0.7  & mIoU \\
     
     \hline
       (4, 8, 16)       & 46.60     & 27.57     & 41.37   & \textbf{33.04}  & \textbf{17.23}  & \textbf{34.61} \\
       (8, 16, 32)      & 48.81     &\textbf{29.60}     & 43.21      & 32.94      & 17.09      & 34.09   \\
       (16, 32, 64)     & \textbf{49.13}  & 29.17  & \textbf{43.53}   & 32.57   & 16.90   & 34.38   \\
       (32, 64, 128)    & 46.72     & 26.44     & 41.66      & 31.58      & 16.70      & 33.57   \\

    \bottomrule
  \end{tabular}}
\end{center}
\end{table}

To further validate the effectiveness of MWFormer for temporal context modeling, we investigate the impact of window sizes, as shown in Table \ref{tab:window_size}. In the MWFormer module, three temporal windows with different sizes are employed to capture multi-scale context within the video feature sequence. In this study, we conduct four experiments with window sizes set to (4, 8, 16), (8, 16, 32), (16, 32, 64), and (32, 64, 128). It can be seen that the proposed method achieves the best performance with window sizes of (16, 32, 64) on Charade-CG, while attaining the best performance with window sizes of (4, 8, 16) on ActivityNet-CG. Accordingly, we adopt (16, 32, 64) for Charades-CG and (4, 8, 16) for ActivityNet-CG in our final experiments. On ActivityNet-CG, smaller window sizes yield superior performance, likely because many annotated moments cover short temporal spans. In such cases, fine-grained local context modeling becomes more effective for precise boundary localization.

\subsubsection{Impact of Masking Ratio in BMFM}

\begin{table}[h]
\begin{center}
\small
  \caption{Impact of mask ratio in BMFM on Novel Composition split.}
  \label{tab:mask_ratio}
  \setlength{\tabcolsep}{1.6mm}{
  \begin{tabular}{ccccccc}
    \toprule
     \multirow{2}{*}{\makecell{\textbf{Masking} \\ \textbf{Ratio}}}  & \multicolumn{3}{c}{\textbf{Charades-CG}}  & \multicolumn{3}{c}{\textbf{ActivityNet-CG}}   \\
      \cmidrule[0.6pt](lr){2-4}
      \cmidrule[0.6pt](lr){5-7}
       & IoU=0.5   & IoU=0.7  & mIoU  & IoU=0.5  & IoU=0.7  & mIoU \\
     
     \hline
       0.1    & \textbf{49.59}   & 28.91   & 43.51   & 30.84   & 15.23   & 32.51  \\
       0.2    & 47.85   & 26.41   & 42.39   & 32.02   & 16.65   & 33.73  \\
       0.3    & 47.36   & 27.75   & 43.13   & 31.95   & \textit{16.76}   & \textit{34.34}  \\
       0.4    & 46.51   & 27.34   & 41.78   & 32.33   & 16.28   & 34.05 \\
       0.5    & 49.13   & \textbf{29.17}   & \textbf{43.53}   & 31.61   & 15.80   & 33.34 \\
       0.6    & 48.98   & 28.99   & 43.20   & 31.93   & 16.43   & 33.60 \\
       0.7    & 47.50   & 28.82   & 42.80   & \textbf{33.04}   & \textbf{17.23}   & \textbf{34.61} \\
       0.8    & 48.17   & 27.89   & 42.38   & 32.70   & 16.66   & 34.22  \\
        
    \bottomrule
  \end{tabular}}
\end{center}
\end{table}

To analyze the effect of different masking ratios on the effectiveness of the proposed bidirectional masked feature modeling (BMFM), we conduct an ablation study by varying the mark ratio from 0.1 to 0.8. For consistency, the same masking ratio is applied to both video and word features. As shown in Table \ref{tab:mask_ratio}, our method achieves the best performance on Charades-CG with a masking ratio of 0.5, whereas a higher masking ratio of 0.7 yields the best results on ActivityNet-CG. This indicates that reconstructing masked video and word features by leveraging information from each other effectively enriches both video and sentence representations. Based on these observations, we adopt masking ratios of 0.5 for Charades-CG and 0.7 for ActivityNet-CG during our model training. 

\subsubsection{Impact of Primitive Prompt Numbers in PPFM on Novel Composition Split.}

\begin{table}[h]
\begin{center}
\small
  \caption{Ablation studies on prompt numbers in PPFM on Novel Composition split.}
  \label{tab:prompt_number}
  \setlength{\tabcolsep}{1.8mm}{
  \begin{tabular}{ccccccc}
    \toprule
     \multirow{2}{*}{\makecell{\textbf{Prompt} \\ \textbf{Number}}}  & \multicolumn{3}{c}{\textbf{Charades-CG}}  & \multicolumn{3}{c}{\textbf{ActivityNet-CG}}   \\
      \cmidrule[0.6pt](lr){2-4}
      \cmidrule[0.6pt](lr){5-7}
       & R1@0.5   & R1@0.7  & mIoU  & R1@0.5  & R1@0.7  & mIoU \\
     
     \hline
       1       & 46.28     & 27.22     & 41.14      & 31.42      & 14.43      & 32.94   \\
       2       & 46.98     & 27.72     & 42.01      & 32.01      & 15.99      & 34.23   \\
       3       & 48.11     & 27.48     & 43.38      & 32.03      & 15.73      & 33.77   \\
       4       & 48.43     & 28.91     & 43.21      & 32.02      & 15.60      & 33.66   \\
       5       & 48.40     & \textbf{29.63}     & 42.99      & 32.26      & 15.90      & 33.62   \\
       6       & \textbf{49.13}     & 29.17     & \textbf{43.53}      & \textbf{33.04}      & \textbf{17.23}      & \textbf{34.61}   \\

    \bottomrule
  \end{tabular}}
\end{center}
\end{table}

In the Primitive Prompt Feature Modulation (PPFM) module introduced in Sec. \ref{sec:prompt}, we enhance the compositional generalization of video temporal grounding by modulating fused video features with individual primitive prompts or their combinations. These primitives, extracted from the sentence, consist of nouns, verbs, adjectives, and adverbs. To assess the effect of the number of prompts, we conduct experiments by randomly selecting between 1 and 6 prompts. As shown in Table \ref{tab:prompt_number}, the best performance is achieved when using 6 prompts, as a larger number of primitives provides richer semantic cues and facilitates alignment with novel compositions during inference. Since the number of prompts depends on sentence length, we set the prompt number to 6 in our experiments.

\subsubsection{Analysis of Loss Functions}

In this study, we analyze the effect of reconstruction loss ($\mathcal{L}_{rec}$) in BMFM and Kullback–Leibler (KL) divergence loss ($\mathcal{L}_{h}$) on the localization performance. As shown in Table \ref{tab:loss}, replacing the cosine distance (CosDist) loss with mean squared error (MSE) for feature reconstruction leads to a clear performance drop, with $mIoU$ decreasing by 4.07\% on Charades-CG and 1.99\% on ActivityNet-CG. This demonstrates that using CosDist as $\mathcal{L}_{rec}$ for both video and sentence features better preserves semantic similarity in the bidirectional masked feature modeling process, thereby yielding superior localization. Furthermore, removing the KL loss from the primitive prompt feature modulation results in additional drops of 0.93\% and 1.01\% on Charades-CG and ActivityNet-CG, respectively. This indicates that $\mathcal{L}_{h}$ encourages each primitive prompt to align with the target moment and thus improves localization accuracy. 

\begin{table}[t]
\begin{center}
\small
  \caption{Comparisons of reconstruction loss and KL Loss on Novel Composition split.}
  \label{tab:loss}
  \setlength{\tabcolsep}{1.8mm}{
  \begin{tabular}{ccccccc}
    \toprule
     \multirow{2}{*}{\makecell{Models}}  & \multicolumn{3}{c}{\textbf{Charades-CG}}  & \multicolumn{3}{c}{\textbf{ActivityNet-CG}}   \\
      \cmidrule[0.6pt](lr){2-4}
      \cmidrule[0.6pt](lr){5-7}
       & R1@0.5   & R1@0.7  & mIoU  & R1@0.5  & R1@0.7  & mIoU \\
     
     \hline
       MSE loss          & 44.83   & 25.51   & 40.46   & 30.75    & 15.61   & 32.62 \\
       w/o KL loss       & 47.73   & 28.44   & 42.60   & 32.27    & 15.56   & 33.60 \\
     \makecell{CosDist loss \\ + KL loss}   & \textbf{49.13}   & \textbf{29.17}   & \textbf{43.53}    & \textbf{33.04}    & \textbf{17.23}   & \textbf{34.61}  \\

    \bottomrule
  \end{tabular}}
\end{center}
\end{table}

\subsubsection{Analysis of Compositional Types}

\begin{table}[h]
\begin{center}
\small
  \caption{Comparison across five composition types in terms of R1@0.5, R1@0.7, and mIoU.}
  \label{tab:composition_types}
  \setlength{\tabcolsep}{1.6mm}{
  \begin{tabular}{c|ccccc}
    \toprule
     \multirow{2}{*}{Metrics}    &\multicolumn{5}{c}{\textbf{Charades-CG}} \\
      & \small{Verb-Noun}   & \small{Adj-Noun}  & \small{Noun-Noun}  & \small{Verb-Adv}  & \small{Prep-Noun} \\
    
    \hline
     Sample Num                       & 1,930  & 23      & 76      & 2,072   & 1,170  \\
    \hline
     

     R1@0.5      & 48.71    & 56.52    & 42.11     & 45.42     & 42.48  \\      
     R1@0.7      & 27.84    & 30.43    & 23.68     & 27.08     & 23.42  \\
     mIoU         & 42.79    & 42.94    & 37.39     & 41.06     & 38.11  \\

     \hline
      \multirow{2}{*}{Metrics}    &\multicolumn{5}{c}{\textbf{ActivityNet-CG}} \\
      & \small{Verb-Noun}   & \small{Adj-Noun}  & \small{Noun-Noun}  & \small{Verb-Adv}  & \small{Prep-Noun}\\
      \hline
      Sample Num                      & 8,704  & 2,257   & 2,399   & 2,456   & 2,598  \\
      \hline
     

      R1@0.5    & 32.43    & 32.74   & 34.39    & 32.65    & 33.60   \\
      R1@0.7    & 16.82    & 18.17   & 19.88    & 15.76    & 17.13   \\
      mIoU       & 34.13    & 35.15   & 36.24    & 34.35    & 34.87   \\
    
    \bottomrule
  \end{tabular}}
\end{center}
\end{table}

To further examine the compositional generalization capability of our method, we evaluate it on five types of word compositions on the Charades-CG and ActivityNet-CG datasets, which include Verb-Noun, Adjective-Noun, Noun-Noun, Verb-Adverb, and Preposition-Noun. Table \ref{tab:composition_types} shows the number of samples corresponding to each compositional type for different datasets, together with their performance in terms of $R@0.5$, $R@0.7$, and $mIoU$. It can be seen that our method has the best performance on the type of Adj-Noun for the Charades-CG dataset, while it has the best performance on the type of Noun-Noun for the ActivityNet-CG dataset. Among these compositional types, the Verb-Noun is the most challenging, as it requires the model to understand their combined semantics for accurate moment localization. Our method achieves competitive or superior results on this type across two datasets, demonstrating that primitive prompts enhance the model's perception of objects and actions in videos.

\subsection{Qualitative Analysis}

\begin{figure*}[h]
\centering
\includegraphics[width=\linewidth]{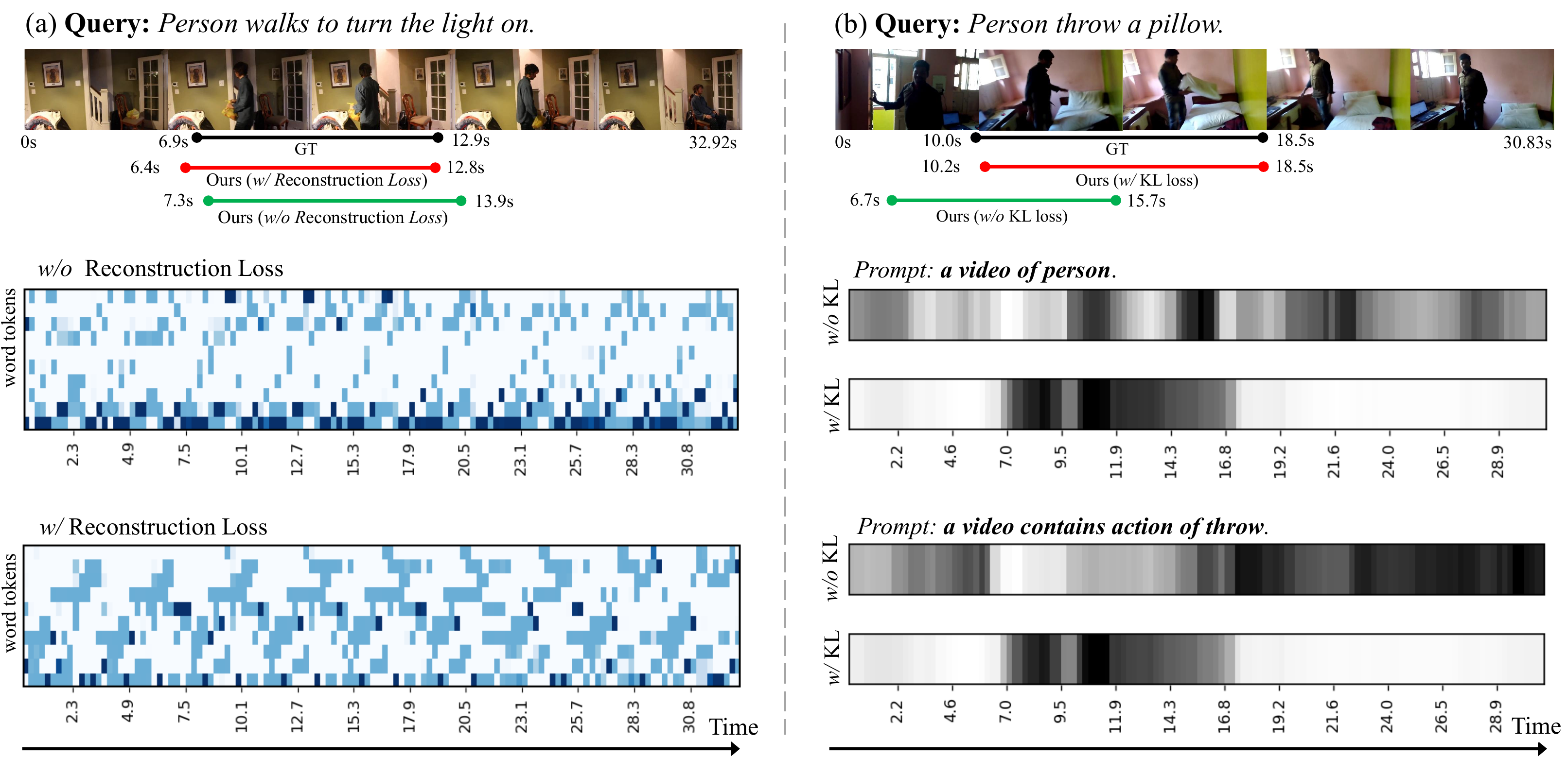}
\caption{Visualization of co-attention weights in BMFM (left) and Gaussian weights in PPFM (right).}
\label{fig:weight}
\vspace{-0.2cm}
\end{figure*}

To further analyze the impact of BMFM and PPFM in our proposed method, we visualize both the co-attention maps and the learnable Gaussian weight distributions. Fig. \ref{fig:weight}(a) presents the co-attention maps without and with BMFM. These maps represent the correlation weights between video features at each temporal position and sentence features across different word positions, where darker colors correspond to higher attention weights. Without BMFM, the attention predominantly concentrates on the final few words of the sentence. In contrast, with BMFM, the attention is more evenly distributed across semantically relevant words in the query. This observation demonstrates that integrating BMFM enhances the model’s ability to capture semantic correspondences between video and text features, thereby improving localization accuracy.

Fig. \ref{fig:weight} (b) illustrates the distributions of the learnable Gaussian weights without and with $\mathcal{L}_{h}$. For the sentence query ``\textit{Person throw a pillow}'', the selected primitive prompts are ``\textit{a video of person}'' and ``\textit{a video contains action of throw}''. It can be observed that the Gaussian weights corresponding to the two primitives exhibit dispersed and inconsistent distributions, with relatively low values around the target moment. In contrast, when $\mathcal{L}_{h}$ is applied, the Gaussian weights for each primitive become more concentrated near the target moment. These results demonstrate the application of $\mathcal{L}_{h}$ can effectively align the textual primitives with the target moment, and improve the model's generalization capability to novel compositions.

\begin{figure*}[!tb]
\centering
\includegraphics[width=\linewidth]{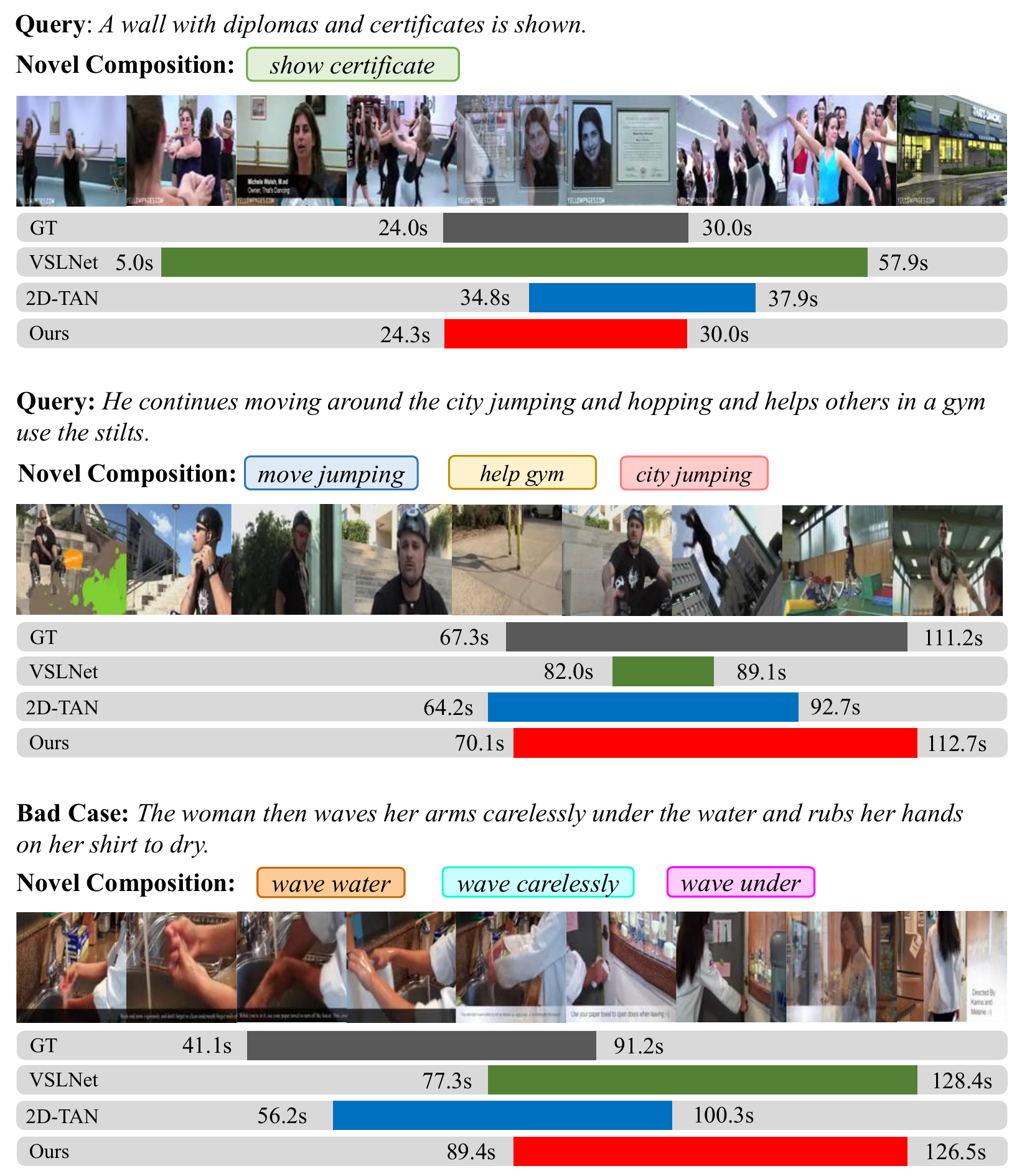}
\caption{Qualitative comparison of results with existing methods.}
\label{fig:visualization}
\vspace{-0.2cm}
\end{figure*} 

Fig. \ref{fig:visualization} presents qualitative comparisons of localization results on three video examples. We compare our method with two publicly available models, VSLNet and 2D-TAN, both trained and evaluated on the compositional temporal grounding datasets. In each example, “GT” denotes the ground truth and ``Our'' refers to the prediction of our method. As shown in Fig. \ref{fig:visualization} (a) and (b), our approach produces results that are more closely aligned with the ground truth, even when the sentence queries involve multiple novel compositions. In contrast, Fig. \ref{fig:visualization} (c) shows a failure case, which may stem from the difficulty of understanding long video content and reasoning over complex scene dynamics. These challenges can hinder the model’s ability to precisely identify the target moment boundaries.

\section{Conclusion}

In this work, we propose a Prompt-guided Primitive Alignment (PPA) framework for compositional temporal grounding, aiming to address the fundamental challenge of generalizing to novel compositions of known concepts in video temporal grounding. To this end, PPA introduces a Primitive Prompt Feature Modulation (PPFM) module that explicitly aligns textual primitives with relevant video regions via learnable Gaussian weighting, and a Bidirectional Masked Feature Modeling (BMFM) strategy that jointly reconstructs masked primitives in both video and text through co-attention interactions. Extensive experiments on Charades-CG and ActivityNet-CG demonstrate that PPA significantly improves compositional generalization, achieving state-of-the-art localization performance on queries with novel compositions and novel words. These results highlight the effectiveness of explicitly modeling primitive-to-moment correspondences and rich bidirectional cross-modal interactions for robust video understanding.

\bibliography{references}

\end{document}